\documentclass[10pt,letterpaper]{article}

\usepackage[margin=1.05in,columnsep=0.20in]{geometry}
\usepackage[T1]{fontenc}
\usepackage{newtxtext}
\usepackage{helvet}
\usepackage{courier}
\usepackage{amsmath}
\usepackage{amssymb}
\usepackage{graphicx}
\usepackage{caption}
\usepackage{float}
\usepackage{placeins}
\usepackage[hyphens]{url}
\usepackage{natbib}

\makeatletter
\renewcommand\normalsize{\@setfontsize\normalsize{10pt}{11pt}}
\renewcommand\small{\@setfontsize\small{9pt}{10pt}}
\renewcommand\footnotesize{\@setfontsize\footnotesize{9pt}{10pt}}
\renewcommand\large{\@setfontsize\large{11pt}{12pt}}
\renewcommand\Large{\@setfontsize\Large{12pt}{14pt}}
\renewcommand\LARGE{\@setfontsize\LARGE{14pt}{16pt}}
\normalsize

\renewcommand\section{\@startsection{section}{1}{\z@}
  {-2.0ex plus -0.5ex minus -0.2ex}
  {3pt plus 2pt minus 1pt}
  {\Large\bfseries\centering}}
\renewcommand\subsection{\@startsection{subsection}{2}{\z@}
  {-2.0ex plus -0.5ex minus -0.2ex}
  {3pt plus 2pt minus 1pt}
  {\large\bfseries\raggedright}}
\renewcommand\subsubsection{\@startsection{subsubsection}{3}{\z@}
  {-6pt plus -2pt minus -1pt}
  {-1em}
  {\normalsize\bfseries}}
\renewcommand\paragraph{\@startsection{paragraph}{4}{\z@}
  {-6pt plus -2pt minus -1pt}
  {-1em}
  {\normalsize\bfseries}}

\newcommand{\affiliations}[1]{\gdef\@affiliations{#1}}
\newcommand{\@affiliations}{}
\renewcommand{\maketitle}{%
  \twocolumn[{%
    \begin{@twocolumnfalse}
      \centering
      \vspace*{0.04in}
      {\LARGE\bfseries \@title\par}
      \vspace{0.10in}
      {\Large\bfseries \@author\par}
      \vspace{0.03in}
      {\normalsize\normalfont \@affiliations\par}
      \vspace{0.13in}
    \end{@twocolumnfalse}%
  }]
  \thispagestyle{empty}%
}
\makeatother

\renewenvironment{abstract}{%
  \centerline{\bfseries Abstract}%
  \vspace{0.5ex}%
  \setlength{\leftmargini}{10pt}%
  \begin{quote}\normalsize%
}{%
  \par\end{quote}\vskip 1ex%
}
\setcitestyle{authoryear,round,semicolon,aysep={}}
\let\cite\citep
\title{QV-PIC: Query-Aware Visual Position-Independent Caching for Efficient RAG Serving}
\author{%
Yilin Liu\textsuperscript{1},
Rui Meng\textsuperscript{2},
Wangze Ni\textsuperscript{1}\textsuperscript{*},
Jianxin Yan\textsuperscript{1},\\
Heng Cao\textsuperscript{3},
Libin Zheng\textsuperscript{4},
Peng Cheng\textsuperscript{5},
Jinfei Liu\textsuperscript{1}%
}
\affiliations{%
\textsuperscript{1}Zhejiang University\\
\textsuperscript{2}Department of Statistics and Data Science, Beijing Normal-Hong Kong Baptist University\\
\textsuperscript{3}Microsoft\\
\textsuperscript{4}Sun Yat-sen University\\
\textsuperscript{5}Tongji University%
}

\begin{document}
\maketitle
\begingroup
\renewcommand{\thefootnote}{\fnsymbol{footnote}}
\footnotetext[1]{Corresponding author.}
\endgroup

\begin{abstract}
Retrieval-Augmented Generation (RAG) repeatedly prefills identical text chunks across queries, incurring redundant computations.
Position-Independent Caching (PIC) mitigates it by reusing precomputed Key-Value (KV) across positions, but its efficiency is constrained by the large volume of text tokens.
Rendering text chunks as images can compress the text into fewer visual tokens, but the rendered-image PIC suffers more severe quality degradation than the text PIC.
This representation-specific gap primarily arises from contextual mismatches across independently compiled caches  and the loss of fine-grained textual evidence during visual compression.
Existing PIC repair methods mainly address the former through selective recomputation, but they incur online computation and cannot recover lost textual details.
We propose QV-PIC, a query-aware dual-resolution PIC reuse framework guided by model-native templates.
Offline, QV-PIC compiles visual caches under the model's native chat-template prefix, improving PIC quality without online recomputation.
Online, it preserves global context with low resolution and restores fine-grained textual evidence within a high-resolution budget by cumulative query relevance scores, retaining the efficiency benefit of visual compression.
Across six tasks, QV-PIC improves average F1 by 21.6 points over vanilla rendered-image PIC, closes the gap to vanilla text PIC, and surpasses optimized text PIC by 2.58 F1 while reducing TTFT by 17.2\%. Relative to full prefill, it cuts TTFT
by 83.8\%.
\end{abstract}

\section{Introduction}
\label{sec:introduction}
Retrieval-Augmented Generation (RAG) augments user queries for Large Language Models (LLMs) with retrieved external documents to support knowledge-intensive tasks~\citep{lewis2020retrieval,guu2020retrieval,karpukhin2020dense,borgeaud2022retro}.
In long-document RAG, identical document chunks are repeatedly prefilled across queries, incurring redundant prefill computation.
Dynamic retrieval rearranges retrieved chunks with different contextual positions, preventing efficient reuse of caches that rely on exact prefix matching or predefined structures~\citep{gim2024prompt,jin2025ragcache,zheng2024sglang}.
Position-Independent Caching (PIC) addresses this limitation by compiling reusable chunks independently and composing their Key-Value (KV) caches at serving time, enabling cross-position reuse of repeated content.
Existing RAG-oriented PIC methods ~\citep{yao2025cacheblend,hu2025epic}  are predominantly text-based.
Although repeated prefill is eliminated, the transmission and computation costs associated with KV caches scale with context length~\citep{kwon2023pagedattention,liu2024cachegen,qin2025mooncake}.

Visual-text compression offers a complementary opportunity to shorten reusable context representations.
By rendering text chunks as compact images, Vision-Language Models (VLMs) can encode multiple textual units into one visual token, increasing per-token information density~\citep{li2025textorpixels,xing2025vist,wei2025deepseekocr}.
We refer to PIC over rendered images as \emph{rendered-image PIC}, in contrast to \emph{text PIC} over the original text chunks.
Glyph~\citep{cheng2025glyph}, for example, achieves 3-4$\times$ token compression while retaining long-context performance comparable to similarly sized text-only LLMs.
This result suggests that rendered images can substantially shorten the reusable KV sequence while preserving useful document information under full prefill, motivating rendered-image PIC reuse across queries for efficient RAG serving.
\begin{figure}[t]
        \centering
        \includegraphics[width=\linewidth]{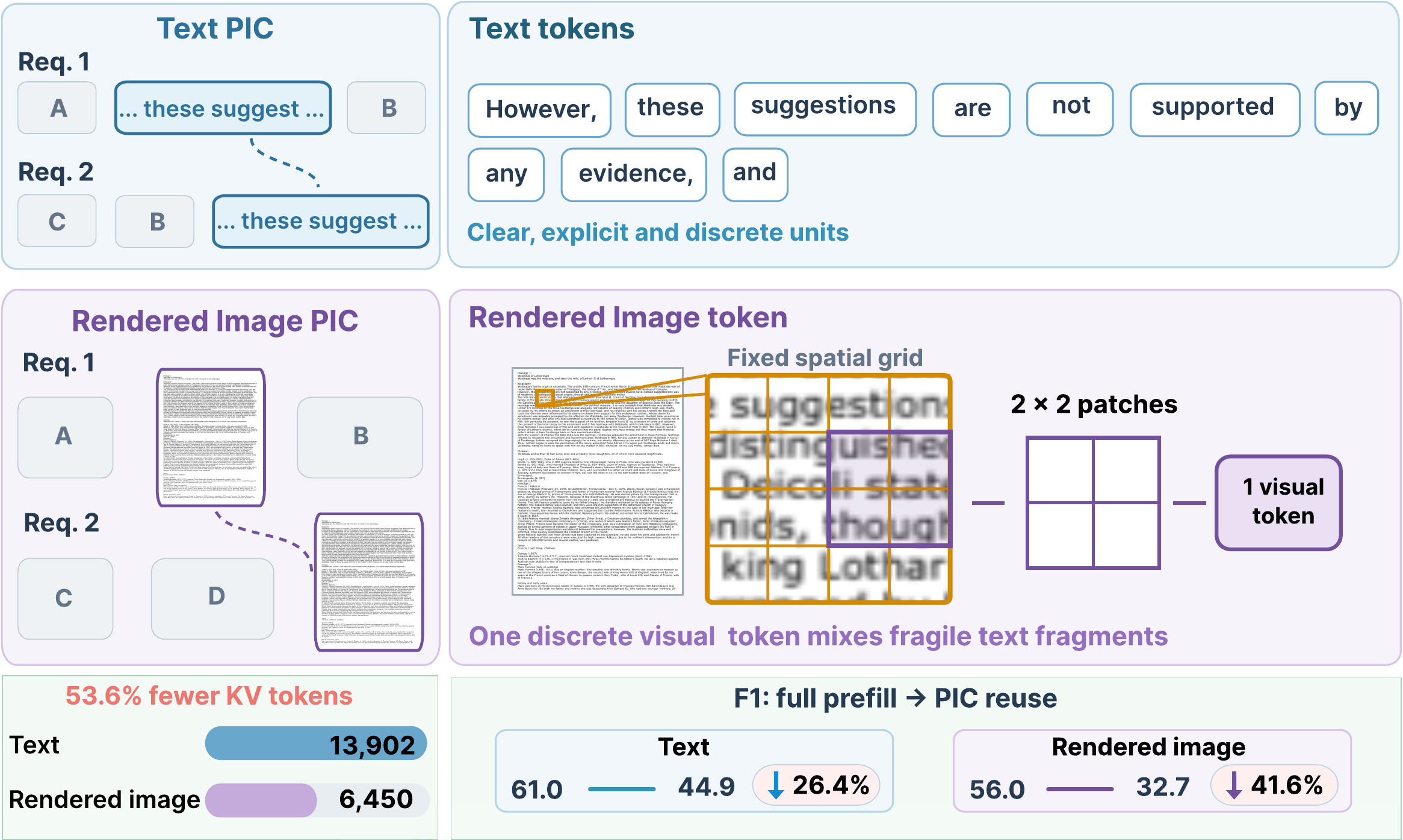}
        \caption{Rendered-image PIC uses fewer KV tokens but incurs a larger full-prefill-to-PIC quality drop than text PIC on Glyph across six LongBench QA tasks at
    72~DPI.}
        \label{fig:motivation}
\end{figure}

This opportunity raises a core question for efficient RAG serving: \textbf{can rendered-image PIC match the reuse quality of text PIC?}
Under identical conditions, we compare text and
rendered-image PIC using Glyph across six LongBench Question-Answering (QA)
tasks.
As shown in Figure ~\ref{fig:motivation}, although rendering substantially reduces token count, rendered-image PIC suffers greater degradation from full prefill than text PIC.
This indicates that rendered images cannot directly inherit the PIC reuse capability of text.
This representation-dependent gap reflects two coupled failure modes.
First, independently compiled caches lack the contextual conditions available during full prefill, causing cache-state mismatch after composition.
Second, text tokenization preserves characters and words as discrete symbols, whereas visual encoding compresses characters, digits, punctuation, and local layout into fewer visual tokens, each aggregating multiple textual units.
Fine-grained answer-bearing evidence may therefore be blurred, conflated, or omitted before the rendered-image KV cache is constructed.
Rendered-image PIC must consequently address both \emph{compilation-context mismatch} and \emph{fine-grained evidence loss}.

Prior PIC repair methods primarily address the first failure mode through selective recomputation or chunk-boundary correction~\citep{yao2025cacheblend,hu2025epic,zhao2025mpic,qin2025vlcache}.
Such methods can refresh context-dependent states after caches are composed, but they introduce additional computation into the online path and operate on an already fixed visual representation.
They cannot reconstruct characters, digits, punctuation, or layout cues that were not preserved during visual encoding.
This limitation is particularly consequential for rendered images, whose answer-bearing information often lies in fine-grained textual details rather than the coarse visual semantics sufficient for natural-image understanding. Consequently,
efficient rendered-image PIC poses two key challenges:

\textbf{Challenge 1: Efficient and stable repair of compilation-context mismatch.}
Prepending disposable prefix during compilation and stripping it afterward can absorb chunk-initial attention sinks without online recomputation.
But arbitrary dummy prefixes induce prefix-dependent cache states, making the repair sensitive to token identity and length.

\textbf{Challenge 2: Efficient restoration of visual-text details.}
Rendering resolution not only controls  textual legibility but also visual-token count.
Uniform low resolution is efficient but risks discarding fine-grained evidence.
High resolution preserves richer textual details but sharply increases tokens,  eroding the efficiency gains of visual-text compression.

In response to these challenges, we propose \textbf{QV-PIC}, a model-native template-conditioned and query-aware dual-resolution
framework for rendered-image PIC, which transforms fixed text chunks into controllable fine-grained units.
\textbf{For Challenge~1}, QV-PIC compiles each rendered image under the model-native chat-template prefix and strips the shared-prefix KV entries before storage. This native template provides the request-invariant prompt-format condition that is present during full prefill, reducing systematic compilation-context mismatch without online recomputation.
\textbf{For Challenge 2}, QV-PIC precompiles low- and high-resolution cache versions for each rendered image.
At serving time, it begins with complete low-resolution context coverage and promotes only a bounded set of query-relevant rendered images to high resolution according to cumulative query relevance.
The two components are complementary: template-conditioned compilation first solidifies the quality foundation of the rendered image PIC, and query-aware dual-resolution allocation then restores query-specific textual details.
The online path requires no rendered-image generation, visual
encoding, or context-side full prefill.
Our contributions are summarized as follows:
\begin{itemize}
\item \textbf{We reveal the significant impact of text and rendered image representations on PIC reuse.}
For identical text content, rendered-image PIC suffers more severe degradation than text PIC, but the former has more potential in quality-latency performance.
\item \textbf{We propose QV-PIC, a template-conditioned and query-aware dual-resolution framework for rendered-image PIC.} It efficiently reduces the compilation-context mismatch and improves fine-grained textual evidence fidelity of rendered-image PIC.
\item \textbf{We demonstrate that QV-PIC   achieves consistent quality-latency
    improvements.}
QV-PIC improves average F1 by 21.6 points over vanilla rendered-image PIC, eliminating its 12.2-point gap to vanilla text PIC and outperforming optimized text PIC by 2.58 points while reducing TTFT by 17.2\%.
Compared with full prefill, it reduces online prefill time by 83.8\%.
\end{itemize}

\section{Background}
\label{sec:background}
\subsection{Position-Independent Caching for Text}
In RAG serving, the same text chunk may be retrieved by different queries with varying prefixes, orders, and contextual positions.
Conventional prefix caching reuses KV caches only when requests share fixed prefixes or predefined layouts.
For example, Prompt Cache~\citep{gim2024prompt} predefines cacheable modules with positions, while RAGCache~\citep{jin2025ragcache} applies tree-based KV retrieval and reuses prefix paths across GPU and memory.
Such prefix-dependent reuse is difficult for dynamic retrieval, where the same chunk may appear in different surrounding contexts.
Text PIC addresses this limitation by independently compiling text chunks into reusable KV caches and linking retrieved caches during online serving.
Cache-Craft~\citep{agarwal2025cachecraft} identifies reusable chunk caches and selectively recomputes context-sensitive states; EPIC~\citep{hu2025epic} formalizes PIC as a compile-and-link framework that recomputes only leading tokens to mitigate attention sinks; TurboRAG~\citep{lu2025turborag} stitches precomputed KV caches with independent attention masks and reordered RoPE positions.
However, these Text PIC methods still incur online overhead from token selection, recomputation, or scheduling.
Although static offline linking can reduce latency, it is constrained by the form and semantics of precompiled prefixes.
More importantly, they fail to shorten text-chunk representations.
Thus, even without full prefill, long-document RAG still requires loading and transferring large text KV caches, limiting PIC reuse efficiency~\citep{liu2024cachegen,qin2025mooncake}.

\subsection{Position-Independent Caching for Images}
Visual-text compression renders text as images, increasing visual-token information density.
Prior work, including Text or Pixels~\citep{li2025textorpixels}, VIST~\citep{xing2025vist}, Glyph~\citep{cheng2025glyph}, DeepSeek-OCR~\citep{wei2025deepseekocr}, and DeepSeek-OCR 2~\citep{wei2026deepseekocr2}, shows its effectiveness for long-context modeling and token compression.
However, compressed visual text remains sensitive to rendering resolution and visual encoding, and OCR readability alone cannot reflect long-range retrieval and reasoning quality~\citep{zhao2025vtcbench}.
Recent work adapts visual processing to task demands:
AgentOCR~\citep{feng2026agentocr} uses segment optical caching to reuse
rendered interaction-history segments, while
AgenticOCR~\citep{wang2026agenticocr} identifies query-relevant regions
and performs OCR on demand.
These methods reduce irrelevant visual input but do not reuse page-level KV caches that can be position-independently assembled across requests.
Multimodal caching methods such as MPIC~\citep{zhao2025mpic} and VLCache~\citep{qin2025vlcache} further reuse visual intermediate states or language-model KV caches with selective recomputation.
However, their fixed ordinary resolution may discard fine-grained textual evidence that later KV repair cannot recover, while recomputation still incurs online overhead.

\section{Methodology}
\label{sec:method}
QV-PIC addresses two sources of degradation in rendered-image PIC:
cache-state mismatch and the loss of fine-grained textual evidence.
Offline, model-native template-conditioned compilation improves
independently compiled KV quality and establishes a reliable reuse
basis. Online, query-aware dual-resolution allocation selectively
restores query-relevant visual details without rerendering or
re-encoding images.
\begin{figure*}[t]
    \centering
    \includegraphics[width=0.95\linewidth]{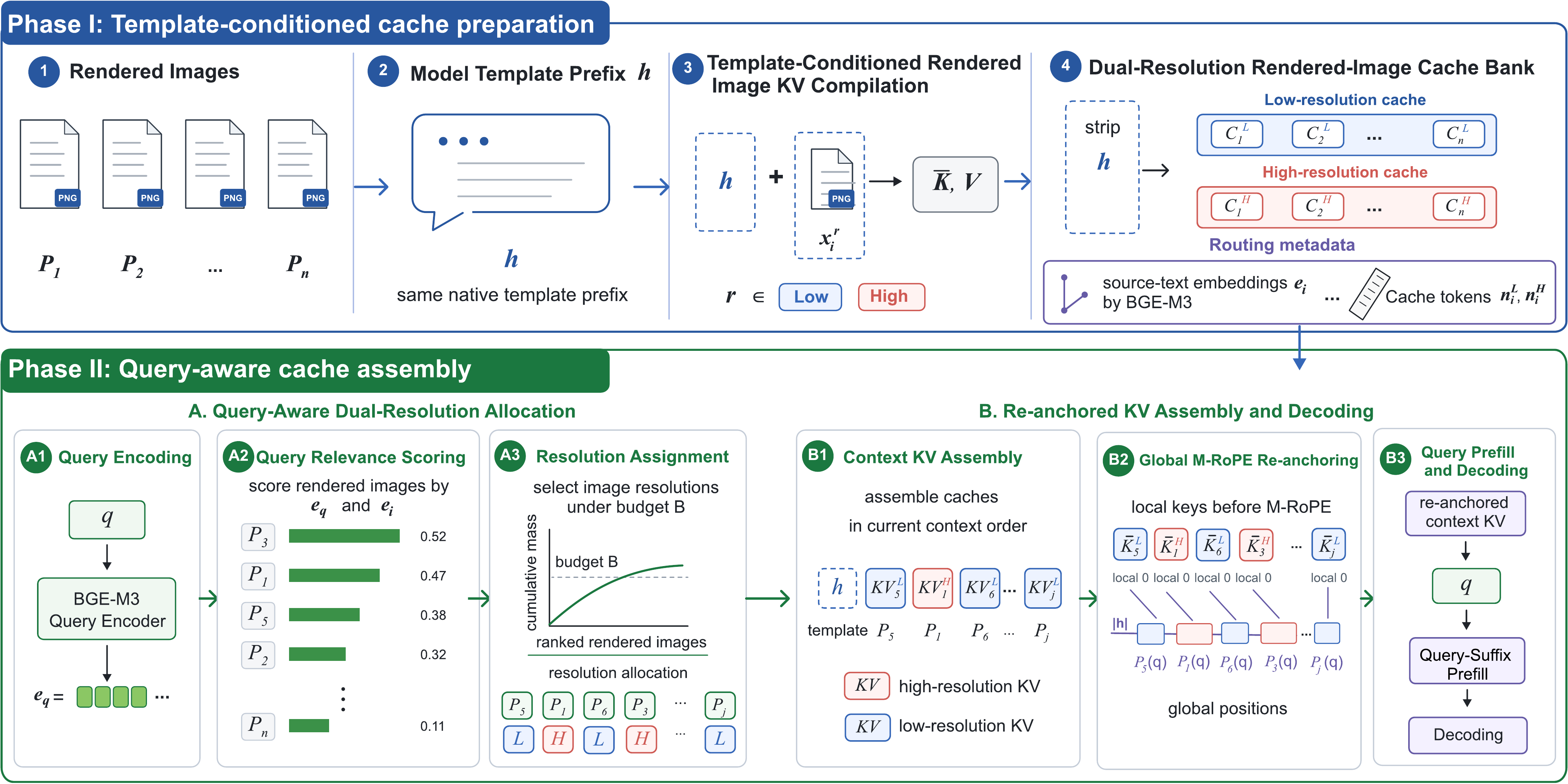}
    \caption{Overview of QV-PIC. Offline, model-native template-conditioned compilation builds low- and high-resolution rendered-image caches and source-text embeddings. Online, query-aware dual-resolution allocation selects one cache version for each rendered image, assembles the selected caches under the current context order, re-anchors M-RoPE positions, and prefills only the query.}
    \label{fig:method}
\end{figure*}

\subsection{Framework Overview}
\label{sec:framework_overview}
Given a query $q$, let
$C(q)=(c_1,\ldots,c_n)$ denote the retrieved chunks in the current context order.
Each chunk $c_i$ is rendered as
$x_i^r=\operatorname{Render}(c_i;r)$ at resolution $r$. As shown in Figure ~\ref{fig:method},
QV-PIC follows a two-phase workflow.

\textbf{Phase I: Template-conditioned cache preparation.}
For each reusable chunk $c_i$, QV-PIC renders low- and high-resolution images and independently compiles one cache per resolution under the model-native chat-template prefix.
It strips the prefix KV entries before storage, but retains the resulting template-conditioned rendered-image KV entries.
The cache bank stores both resolution variants, token-count metadata, source-order metadata, and a source-text embedding for later routing.

\textbf{Phase II: Query-aware cache assembly.}
At serving time, QV-PIC scores the retrieved chunks against the query and promotes at most $B$ query-relevant rendered images to high resolution.
All retrieved chunks remain in the context, and exactly one cache version is activated for each rendered image.
The relevance ranking is used only for resolution allocation; cache assembly follows the current context order of the RAG request.
After M-RoPE re-anchoring, the assembled cache is supplied as past KV, so the VLM only computes query prefill and answer generation online.

\subsection{Model-Native Template-Conditioned Compilation}
\label{sec:composable_cache}
In full prefill, the rendered-image context is processed under the model-native chat-template prefix, which provides the request-invariant prompt-format condition of the VLM's multimodal serving interface.
Prefix-free compilation removes this condition, while dummy prefixes replace it with arbitrary tokens whose effect varies with token identity and length.
QV-PIC instead uses the chat-template prefix as the compile-time condition and strips only its KV entries before storage.
This reduces prompt-format mismatches in independently compiled rendered-image caches.

Let $M$ be the VLM and $h$ its model-native chat-template prefix.
For a rendered image $x_i^r$, QV-PIC constructs
\begin{equation}
    \mathcal C_i^r
    =
    \operatorname{Strip}_{h}
    \left(
        \operatorname{KV}_{M}([h;x_i^r])
    \right),
\end{equation}
where $\operatorname{Strip}_{h}$ removes the KV entries corresponding to $h$.
Although these prefix entries are discarded, the retained rendered-image entries are still computed under the native template condition.
At serving time, QV-PIC adds only one shared prefix cache,
$\mathcal C_h=\operatorname{KV}_{M}(h)$.

Since each rendered-image cache is compiled independently, its keys are stored before M-RoPE rotation~\citep{su2024roformer,wang2024qwen2vl}.
After the current context order and resolution assignment are fixed, QV-PIC derives the request positions $\mathbf P_i(q)$ for each activated cache and applies
\begin{equation}
\mathbf K_{\ell,i}^{r}
=
\mathcal R_{\ell}
\left(
\bar{\mathbf K}_{\ell,i}^{r},
\mathbf P_i(q)
\right).
\end{equation}
where $\bar{\mathbf K}_{\ell,i}^{r}$ is the unrotated key at layer $\ell$, and $\mathcal R_{\ell}$ is the model-native M-RoPE operator.
Values are position-independent and are stitched in the same current context order.
Template conditioning improves the independently compiled rendered-image KV entries, while M-RoPE re-anchoring places them at their request positions.
Thus, the two operations are complementary.

\subsection{Query-Aware Dual-Resolution Allocation}
\label{sec:adaptive_assembly}

Uniform low resolution reduces visual-token and KV costs, but may weaken characters, numbers, and local textual evidence.
Uniform high resolution preserves more detail, but increases the active KV size for every rendered image.
QV-PIC therefore precompiles both versions and activates high resolution only for query-relevant rendered images:
\begin{equation}
    \mathcal B_i=\{\mathcal C_i^L,\mathcal C_i^H\},
\end{equation}
where $\mathcal C_i^L$ and $\mathcal C_i^H$ denote the low- and high-resolution caches of rendered image $i$.

\paragraph{Query-relevance scoring.}
QV-PIC uses a frozen BGE-M3 encoder~\citep{chen-etal-2024-m3} $E(\cdot)$ to embed the source text of each rendered image offline and the query online:
\begin{equation}
    \mathbf e_i=\frac{E(c_i)}{\|E(c_i)\|_2},
    \qquad
    \mathbf e_q=\frac{E(q)}{\|E(q)\|_2}.
\end{equation}
The relevance score is cosine similarity:
\begin{equation}
    \tilde{s}_i=\mathbf e_q^\top \mathbf e_i,
    \qquad
    s_i=[\tilde{s}_i]_+=\max(\tilde{s}_i,0).
\end{equation}
Let $\pi(q)$ sort the retrieved chunks by $\tilde{s}_i$ in descending order.
This ranking is used only to choose high-resolution caches.
When $\sum_i s_i>0$, QV-PIC selects the smallest top-ranked set whose cumulative positive relevance reaches threshold $\alpha$, capped by budget $B$:
\begin{equation}
    k^\star
    =
    \min
    \left(
        B,\,
        \min\left\{
            k:
            \frac{\sum_{j=1}^{k}s_{\pi_j}}
                 {\sum_{i=1}^{n}s_i}
            \geq \alpha
        \right\}
    \right).
\end{equation}
If all scores are non-positive, QV-PIC selects the highest-scoring chunk as a deterministic fallback.
The promoted set and resolution assignment are
\begin{equation}
    \mathcal S(q)=\{\pi_1,\ldots,\pi_{k^\star}\},
    \qquad
    r_i(q)=
    \begin{cases}
        H, & i\in\mathcal S(q),\\
        L, & \text{otherwise}.
    \end{cases}
\end{equation}
Here, $\tilde{s}_i$ measures query relevance, whereas $r_i(q)$ denotes the assigned resolution.

\paragraph{Online assembly and cost.}
After resolution assignment, QV-PIC activates
$\{\mathcal C_i^{r_i(q)}\}_{i=1}^{n}$
and assembles them under the current context order.
If $n_i^L$ and $n_i^H$ denote the low- and high-resolution token counts of rendered image $i$, the active rendered-image prefix length is
\begin{equation}
    N(q)
    =
    \sum_{i=1}^{n} n_i^L
    +
    \sum_{i\in\mathcal S(q)}
    \left(n_i^H-n_i^L\right),
    \qquad
    |\mathcal S(q)|\leq B.
\end{equation}
Thus, high-resolution overhead is paid only for promoted rendered images.
Online routing requires query encoding, similarity scoring, and ranking:
\begin{equation}
    T_{\mathrm{route}}
    =
    T_E(q)+O(nd)+O(n\log n),
\end{equation}
where $d$ is the embedding dimension.
Rendering, visual encoding, and rendered-image KV compilation remain offline.

\section{Experiments}
\label{sec:experiments}
In this section, we conduct experiments to evaluate QV-PIC by addressing the following questions:

\textbf{Q1}: Can model-native template-conditioned compilation improve the reuse quality of independently compiled rendered-image caches?

\textbf{Q2}: Under template-conditioned compilation, how do the F1 and TTFT of rendered-image PIC change with uniform DPI scaling?

\textbf{Q3}: Can QV-PIC achieve higher average F1 with lower average TTFT than uniform 120-DPI rendered-image PIC and text PIC?

\textbf{Q4}: How well does QV-PIC generalize beyond Glyph?

\subsection{Experimental Configuration}
\label{sec:experimental_configuration}

\subsubsection{Implementation}
We implement all methods in a unified Hugging Face-PyTorch inference framework
and evaluate them on a server equipped with eight NVIDIA A800 80~GB GPUs. All methods using the same
backbone share identical configurations.
Following the rendering protocol of Glyph ~\citep{cheng2025glyph}, we fix the rendering canvas size, margins, font, and line spacing while varying only DPI. We extend Glyph's 72/96/120-DPI range to 144 and 168~DPI at 24-DPI intervals. QV-PIC uses 72/120~DPI as its dual-resolution configuration: 72~DPI preserves full-context coverage at low cost, whereas 120~DPI provides a clear average quality gain without the larger token and latency costs of 144 and 168~DPI. For query-aware dual-resolution allocation, we rank rendered images by relevance and select the smallest top-ranked set whose cumulative normalized positive relevance reaches $\alpha=0.65$, capped at $B=4$ high-resolution rendered images. Owing to the compute budget, NarrativeQA is evaluated at 72, 96, and 120~DPI, whereas the other tasks use the complete DPI sweep.

\subsubsection{Model Selection}
For \textbf{Q1-Q3}, we use Glyph 9B as the primary model. Glyph receives rendered-text-specific adaptation through continual pretraining on rendered long-text data and OCR-aware SFT/RL. This specialization reduces confounding from basic rendered-text recognition, allowing Q1-Q3 to focus on cache compilation and resolution allocation.
For \textbf{Q4}, we evaluate two technically compatible
general-purpose VLMs of comparable scale:
GLM-4.1V-9B-Thinking~\cite{glmvteam2025glm} and
LLaVA-OneVision-2-8B-Instruct~\cite{an2026llavaonevision2}.
Both support multi-image inputs and provide OCR and
document-understanding capabilities, but neither has undergone
rendered-text-specific adaptation. GLM-4.1V provides a
related-family setting because Glyph is initialized from
GLM-4.1V-9B-Base, whereas LLaVA-OneVision-2 uses a
different vision encoder, language backbone, and training recipe,
providing a cross-family setting. These models are used as
conservative transfer probes.
Positive
results on them would indicate that QV-PIC does not rely entirely
on Glyph's rendered-text-specific training. Meanwhile, dedicated
rendered-text adaptation may provide additional quality headroom
for QV-PIC on future compatible backbones.

\subsubsection{Baselines}
For \textbf{Q1}, we compare a prefix-free baseline with three cache-state repair strategies: dummy-prefix conditioning using $k\in\{2,4,8,16\}$ repetitions of the placeholder token \texttt{x}, model-native template conditioning, and an efficient recomputation method EPIC-2/4 without token selection. The $k=4$ dummy prefix matches the native chat-template length, while the remaining lengths test sensitivity to arbitrary prefix length.
For \textbf{Q2}, we compare full prefill and template-conditioned PIC for both text and rendered-image inputs across DPI settings. This separates representation quality from PIC degradation and evaluates the quality and latency effects of uniform DPI scaling.
For \textbf{Q3}, we compare QV-PIC with template-conditioned text PIC, uniform 72- and 120-DPI rendered-image PIC, and QV-PIC without template conditioning. This isolates the contribution of dual-resolution allocation and its complementarity with template conditioning.
\textbf{Q4} repeats the same within-backbone comparison on two additional VLMs, measuring generalization relative to each model's own PIC baseline.

\subsubsection{Datasets}
For \textbf{Q1-Q4}, we select six long-context question-answering (QA) tasks from
LongBench~\citep{bai-etal-2024-longbench}. 2WikiMQA, HotpotQA, and MuSiQue
cover multi-document, multi-hop evidence aggregation.
MultiFieldQA-en and
NarrativeQA evaluate evidence localization and holistic understanding within long
single documents, while TriviaQA focuses on factoid question answering. Together,
these tasks span single- and multi-document contexts, localized and distributed
evidence, and direct retrieval and multi-hop reasoning, providing complementary
tests of the composition and reuse of independently compiled rendered-image caches.
We use all 1,150 examples in LongBench evaluation subsets.
MultiFieldQA-en contains 150 examples, and each of the other five tasks contains
200. Results are first averaged within
each task and then equally averaged across tasks.

\subsubsection{Metrics}
QV-PIC is evaluated  by answer quality, online latency, and token size.
Answer quality is measured by official LongBench token-overlap F1
using one deterministic run per example. TTFT, averaged over three
runs, is measured from a CUDA synchronization immediately before
each online request to first-token logits.
For full prefill, TTFT includes visual encoding when applicable, full-context prefill, and first-token computation.
For PIC, TTFT includes CPU-to-GPU KV transfer and materialization, cache composition, global positional re-anchoring, query-suffix prefill, and first-token computation.
QV-PIC additionally includes BGE-M3 query encoding, relevance scoring, ranking, and resolution assignment.

\subsection{Q1: Effectiveness of Model-Native Template-Conditioned Compilation}
\label{sec:q1_prefix_conditioning}
To answer Q1, we compare prefix-free compilation, dummy-prefix compilation with $k \in \{2,4,8,16\}$, and model-native template-conditioned compilation.
For the latter two, the prefix is prepended during offline compilation and their KV entries are then discarded, retaining only the prefix-conditioned KV.
Rendered-image experiments use 72 and 120 DPI. Since the native chat-template prefix has four tokens, dummy-prefix-4 serves as a length-matched control.
We also compare EPIC-2/4, which recomputes the first two or four chunk tokens online.
Figures~3 and~4 report the six-task average F1.
\begin{figure}[H]
    \centering
    \includegraphics[width=0.93\linewidth]{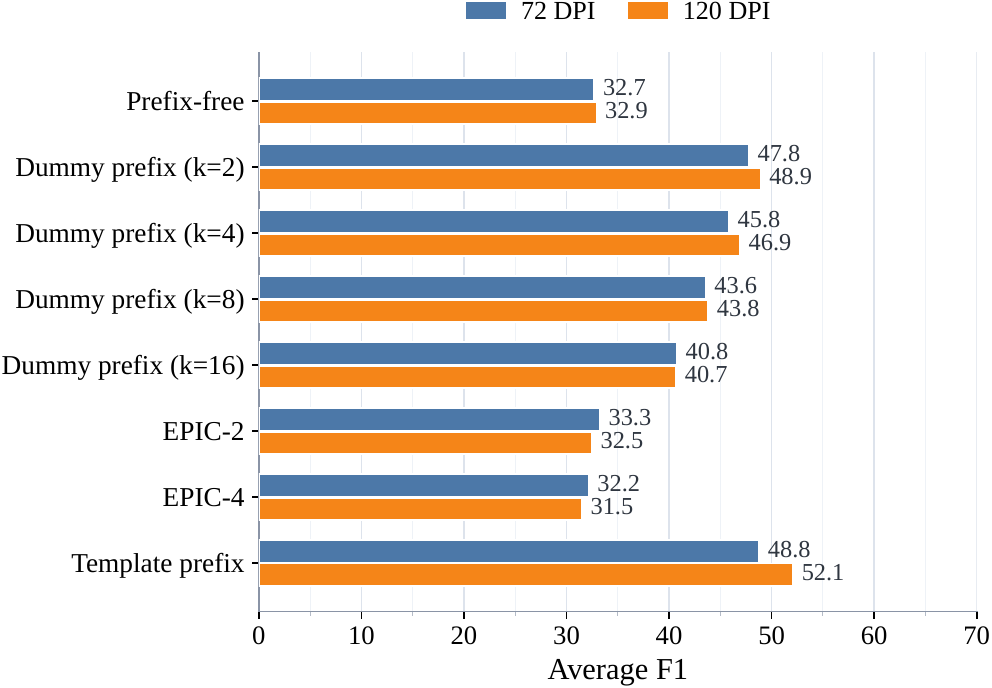}
        \caption{Six-task average F1 of rendered-image PIC under different cache-compilation and repair settings at 72 and 120~DPI.}
        \label{fig:q1_image_header_mean}
\end{figure}
\paragraph{Prefix-conditioned compilation outperforms resolution scaling and  leading-token recomputation.}
Prefix-free rendered-image PIC obtains average F1 scores of 32.7 and 32.9 at 72 dpi and 120 dpi, respectively, which are 12.2 and 12.0 points lower than prefix-free text PIC. Increasing DPI provides almost no improvement.
EPIC-2/4 achieves F1 scores of 31.5 and 33.3, remaining comparable to prefix-free Rendered-Image PIC. In contrast, the dummy prefix $k=2$ improves  F1 to 47.8 and 48.9 at the two resolutions, indicating that conditioning each chunk during offline compilation is more effective than recomputing a few leading tokens of each chunk online.
However, as the dummy-prefix length increases from 2 to 16, the average F1 of both image and text drops substantially, indicating its sensitivity to  arbitrary prefix length.
Template-conditioned compilation achieves average F1 scores of 48.8, 52.1, and 51.7 for 72-DPI rendered images, 120-DPI rendered images, and text, respectively. It
outperforms the length-matched dummy prefix by 3.0, 5.2, and 5.1 points.
This confirms that the gain comes from alignment with the model's learned input interface rather than the mere presence or length of prefix conditioning.
\paragraph{Answer to Q1.}\textbf{Model-native template conditioning provides the highest PIC quality across resolutions and modalities.}
At 120~DPI, it raises the rendered-image PIC from 32.9 to 52.1 F1, converting its original 12.0-point gap from text PIC into a 0.4-point advantage. Therefore, model-native template conditioning constructs higher-quality reusable caches entirely offline, without online  recomputation or tuning an arbitrary dummy-prefix length.
\begin{figure}[H]
        \centering
\includegraphics[width=0.85\linewidth]{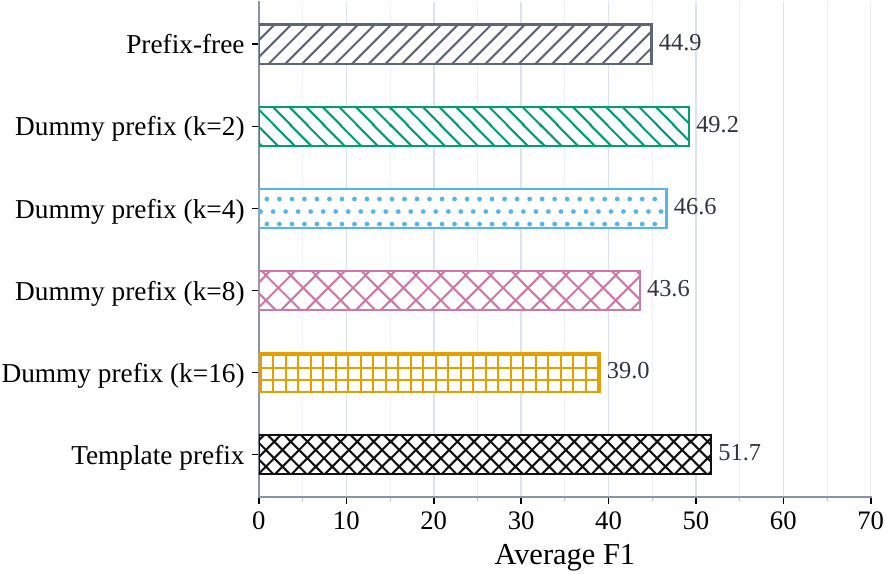}
        \caption{Six-task average F1 of text PIC under different cache-compilation settings.}
        \label{fig:q1_text_header_mean}
\end{figure}

\subsection{Q2: Effects of Uniform DPI Scaling on Quality and Latency} \label{sec:q2_single_dpi}
Q1 shows that DPI alone cannot repair independently compiled caches, whereas template conditioning establishes a reliable reuse basis and allows higher resolution to deliver further average F1 gains. Q2 therefore examines how uniform DPI scaling affects the F1 and TTFT of rendered-image PIC relative to text PIC and full prefill.
\begin{figure}[t]
    \centering
    \includegraphics[width=0.45\textwidth]{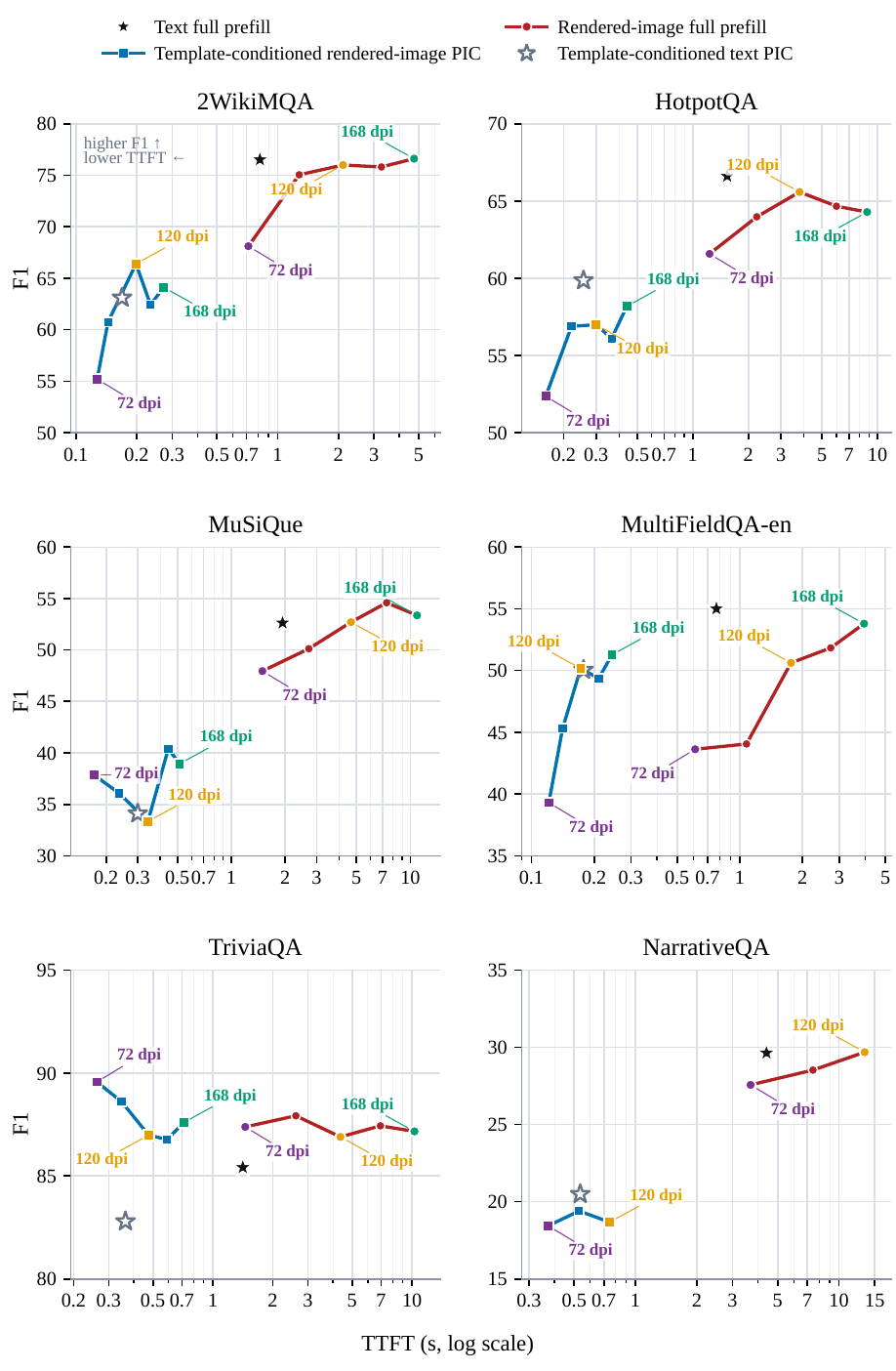}
    \caption{Per-task F1-TTFT comparison of full prefill and template-conditioned PIC for text and rendered image across DPI settings. Image points are connected in ascending DPI order, and TTFT is shown on a logarithmic scale.}
   \label{fig:q2_uniform_resolution_tradeoff}
\end{figure}
\paragraph{Uniform DPI scaling yields unstable F1 changes while TTFT increases consistently.}
As shown in Figure~\ref{fig:q2_uniform_resolution_tradeoff}, the best rendered-image full-prefill configurations approach or match text full prefill across the six tasks, confirming the quality potential of rendered-image inputs.
Template-conditioned rendered-image PIC improves from an average F1 of 48.8 at 72~DPI to 52.1 at 120~DPI, and at least one DPI setting reaches or exceeds text PIC on four tasks.
However, the per-task F1 gains are non-monotonic. Increasing DPI may improve, preserve, or reduce F1, showing that additional visual detail does not reliably translate into higher reuse quality. In contrast, TTFT increases consistently as visual tokens and KV caches grow. Although increasing DPI provides more visual detail,
it also adds visual tokens whose additional detail is not
consistently useful to the current query.
Moreover, even at 120~DPI, rendered-image PIC remains roughly an order of magnitude faster than rendered-image full prefill on most tasks.

\paragraph{Answer to Q2.}
\textbf{Template conditioning enables moderate DPI increases to improve rendered-image PIC quality while retaining substantial full-prefill speedups.}
However, the gains become limited or unstable whereas TTFT increases consistently, motivating selective high-resolution allocation to query-relevant rendered images.

\subsection{Q3: Joint F1-TTFT Improvement via Query-Aware Dual-Resolution Allocation}
\label{sec:q3_adaptive_resolution}
\begin{figure}[H]
        \centering
       \includegraphics[width=\linewidth]{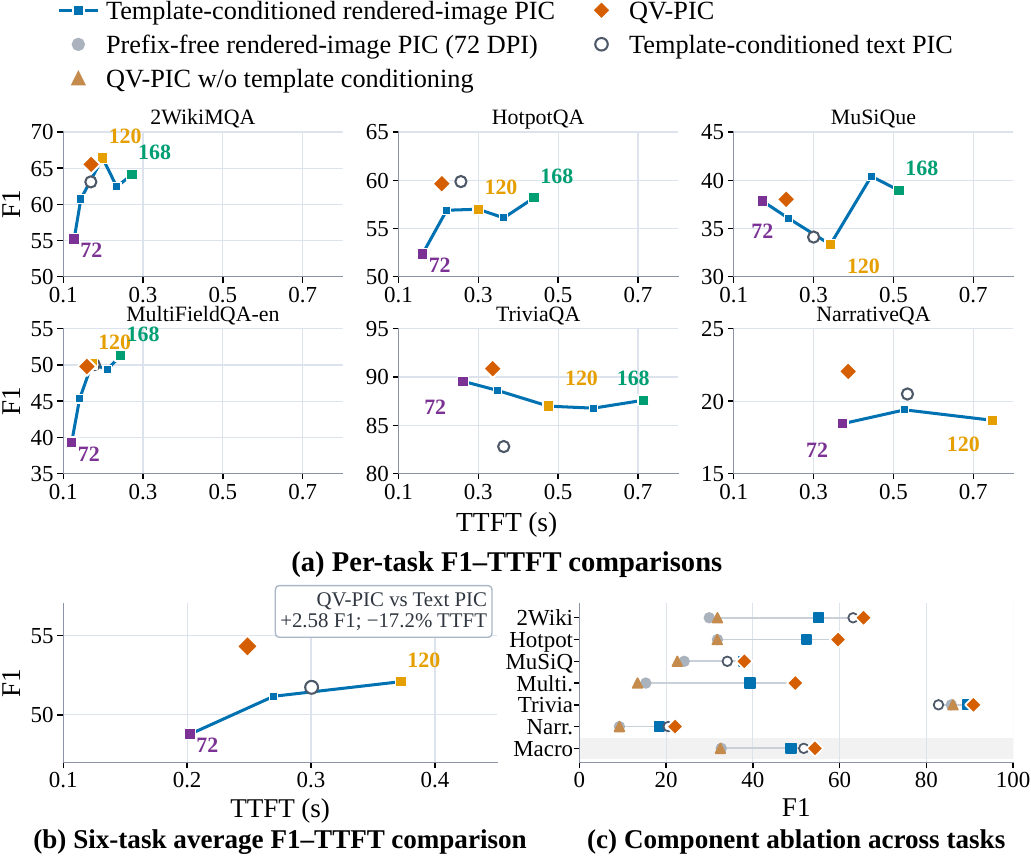}
        \caption{F1-TTFT comparisons on different PIC methods and component ablation of QV-PIC.}   \label{fig:q3_adaptive_resolution_tradeoff}
\end{figure}

Q3 examines whether allocating a bounded high-resolution budget to query-relevant rendered images can improve the overall performance of rendered-image PIC reuse. QV-PIC retains most rendered-image caches at 72~DPI and selects the most relevant ones with 120 DPI. Additionally, we remove template conditioning to evaluate its synergy with query-aware dual-resolution allocation.
\paragraph{Query-aware dual-resolution allocation improves F1 without uniform high-resolution overhead.}
As shown in Figure ~\ref{fig:q3_adaptive_resolution_tradeoff}, QV-PIC simultaneously improves F1 and reduces TTFT over uniform 120-DPI rendered-image PIC on HotpotQA, MuSiQue, TriviaQA, and NarrativeQA, indicating that enhancing only query-relevant images preserves useful visual-detail gains while avoiding unnecessary visual overhead on irrelevant pages.
Compared with text PIC, it improves F1 and reduces TTFT on MuSiQue, TriviaQA, and NarrativeQA, achieves higher F1 at comparable TTFT on 2WikiMQA, and maintains similar F1 at lower TTFT on HotpotQA and MultiFieldQA-en.
Moreover, template conditioning raises the average F1 of prefix-free 72-DPI PIC from 32.7 to 48.8, whereas dual-resolution allocation alone reaches only 32.5. Combining both components increases the average F1 to 54.3, with gains on all tasks, finally surpassing the 51.7 F1 of text PIC. Template conditioning therefore establishes a reliable KV basis, while query-aware dual-resolution allocation provides query-relevant fine-grained evidence.
\paragraph{Answer to Q3.} \textbf{QV-PIC preserves the visual compression advantage while achieving better overall F1.} Compared with  text PIC and template-conditioned 120-DPI rendered-image PIC, QV-PIC attains higher average F1 with lower average TTFT. The ablation further confirms that the gains arise from the synergism of template-conditioned compilation and query-aware dual-resolution allocation.

\subsection{Q4: Cross-Model Generalization of QV-PIC} \label{sec:q4}
Q4 examines whether QV-PIC remains effective on general-purpose VLMs GLM-4.1V and LLaVA-OneVision-2.
GLM-4.1V provides a related-family setting, whereas LLaVA-OneVision-2 provides a cross-family test. We compare prefix-free 72 DPI rendered-image PIC, template-conditioned rendered-image PIC at 72 and 120 DPI, template-conditioned text PIC, and QV-PIC.

\begin{figure}[H]
        \centering
       \includegraphics[width=\linewidth]{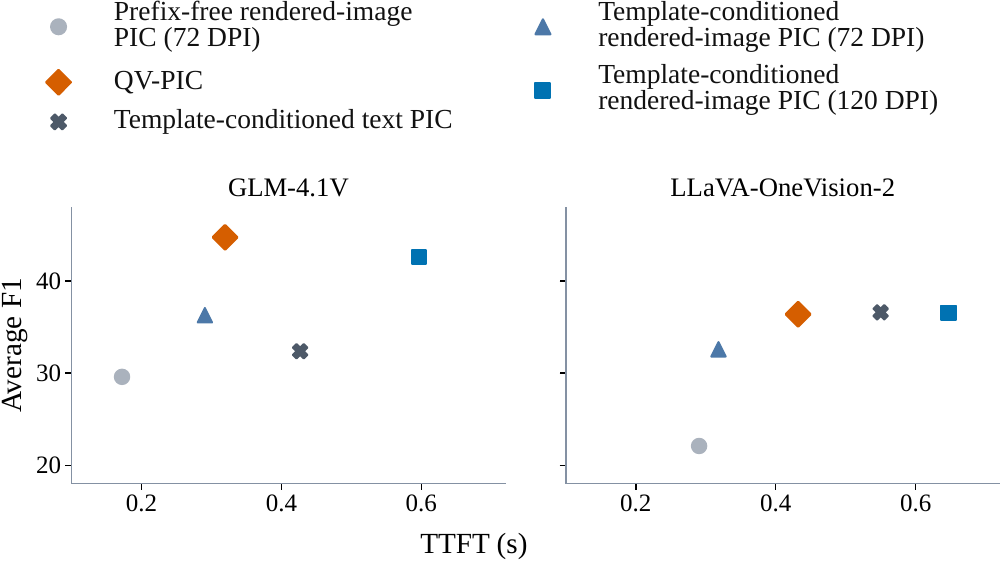}
       \caption{Cross-model six-task average F1-TTFT comparison of QV-PIC.}
        \label{fig:q4_cross_model}
\end{figure}

\paragraph{QV-PIC consistently strengthens rendered-image PIC.}
As shown in Figure~\ref{fig:q4_cross_model}, template-conditioned 72-DPI rendered-image PIC substantially improves average F1 over prefix-free compilation on both models.
Uniform 120-DPI compilation further improves F1 on both models, confirming that
template-conditioned compilation is not confined to Glyph's rendered-text-specific adaptation.
On GLM-4.1V, QV-PIC achieves the highest average F1 while requiring lower TTFT than both uniform 120-DPI rendered-image PIC and template-conditioned text PIC. On LLaVA-OneVision-2, QV-PIC substantially improves over the uniform 72-DPI configuration and nearly matches the highest average F1 obtained by uniform 120-DPI rendered-image PIC and text PIC, while requiring markedly lower TTFT than either.

\paragraph{Answer to Q4.}
\textbf{QV-PIC shows promising generalization beyond the primary Glyph model.}
Across both related- and cross-family general-purpose VLMs, it either achieves the highest F1 with lower TTFT or retains a near-best F1 at a lower TTFT. Thus, its gains are not limited to Glyph's rendered-text specialization, although the magnitude of the gain depends on the underlying VLM.

\section{Conclusion}
Regarding the reuse-quality degradation caused by cache-state mismatch and fine-grained evidence loss in rendered-image PIC, we propose QV-PIC, a query-aware visual PIC framework for efficient RAG serving.
It combines model-native template-conditioned cache compilation with query-aware dual-resolution cache assembly to reduce compilation-context mismatch and preserve query-relevant textual evidence.
Across six LongBench QA tasks, QV-PIC improves rendered-image PIC by 21.6 F1 points and surpasses optimized text PIC and uniform 120-DPI rendered-image PIC with lower TTFT.
Compared with full prefill, it reduces online prefill time by 83.8\%, enabling fast and accurate long-document RAG with reusable visual-text caches.

\FloatBarrier
\bibliography{references}

\end{document}